\RequirePackage{snapshot}
 \documentclass[tablecaption=bottom,wcp]{jmlr} % W&CP article

\usepackage{booktabs}
\usepackage[load-configurations=version-1]{siunitx} % newer version
\theorembodyfont{\upshape}
\theoremheaderfont{\scshape}
\theorempostheader{:}
\theoremsep{\newline}

\jmlrproceedings{AABI 2023}{5th Symposium on Advances in Approximate Bayesian Inference, 2023}

\title[Variational Prediction]{Variational Prediction}

\author{\Name{Alexander A. Alemi}
\Email{alemi@google.com}\and\Name{Ben Poole} \Email{pooleb@google.com}\\
  \addr Google Research}
\usepackage{tikz}
\usepackage{mycustom}

\begin{document}

\maketitle

\begin{abstract}
%Everyone knows that finding the Bayesian posterior predictive distribution requires marginalizing out the posterior.  What this paper presupposes is\dots maybe it doesn't? 
%[TODO: we need an actual abstract]
% \fixme{rewrite}

% Bayesian inference comes with benefits over maximum likelihood, but it also comes with costs.  It is typically intractable not only to compute the posterior, but also to marginalize that posterior to form the posterior predictive distribution.  Here we discuss a technique for directly learning a variational approximate posterior predictive distribution by means of a variational bound that can provide good predictive distributions without any test time marginalization costs.  We demonstrate this Variational Prediction technique on a illustrative toy example.

Bayesian inference offers benefits over maximum likelihood, but it also comes with computational costs. Computing the posterior is typically intractable, as is marginalizing that posterior to form the posterior predictive distribution. In this paper, we present \emph{variational prediction}, a technique for directly learning a variational approximation to the posterior predictive distribution using a variational bound. 
This approach can provide good predictive distributions without test time marginalization costs. We demonstrate Variational Prediction on an illustrative toy example.
\end{abstract}

% Keywords may be removed
%\begin{keywords}
%List of keywords
%\end{keywords}

%\documentclass[../main.tex]{subfiles}
% \graphicspath{{\subfix{../images/}}}
%\begin{document}

\section{Introduction}

The promise of Bayesian inference is that it can provide accurate predictions by leveraging prior knowledge about the world and its mechanisms.
Unfortunately, computing these predictions is costly, as it requires integrating over all possible parameter settings.
%More formally, we start 

Given a parametric statistical \emph{model} $p(x|\theta)$
and \emph{prior} $p(\theta)$,
the \emph{posterior distribution} over the parameters of our
model is given by:
\footnote{We use $p(D|\theta)$ as shorthand for the likelihood over a dataset, $D \equiv \{ x_1, x_2, \ldots, x_N \}$, $p(D|\theta) \equiv \prod_i p(x_i|\theta)$.}
\begin{equation}
    p(\theta|D) = \frac{p(D|\theta)p(\theta)}{p(D)} \quad \textrm{with} \quad  p(D) = \int d\theta\, p(\theta) p(D|\theta).
\end{equation}
We marginalize out this posterior to compute the 
\emph{posterior predictive} distribution:
\begin{equation}
    p(x|D) = \int d\theta\, p(x|\theta)p(\theta|D),
\end{equation}
which is the optimal predictive distribution, on average, in the well-specified case~\citep{aitchison1975goodness}. 
% and whose marginalization can provide benefits beyond~\citep{caseforbayes}. 

% It is this last step of marginalization that is thought to provide many of the observed benefits of the Bayesian approach~\citep{caseforbayes}. 

% The Bayesian posterior predictive distribution is optimal on average in well-specified setting~\citep{aitchison1975goodness}.
%This predictive distribution 
%Bayesian inference and the corresponding predictions made through the posterior predictive distribution have led to advances in numerous fields
%[TODO: add citations for how amazing Bayesian inference is].
% 
%However, there are two major challenges:
% The marginalization of the \emph{predictive} approaches can provide benefits~\citep{caseforbayes} over the point-like \emph{estimative} approaches.
Unfortunately, both computing the posterior distribution $p(\theta|D)$ and marginalizing this posterior to form the posterior predictive $p(x|D)$ are often intractable.
% Unfortunately, there are two typically intractable problems that need to be tackled to perform Bayesian inference.
%Unfortunately there are typically some problems.
%in applying the tools of Bayesian inference to large-scale non-parametric models in the modern era:
% First, exactly computing the posterior distribution, $p(\theta | D)$ over parameters is intractable in all but the simplest cases. Second, marginalizing out the parameters to form the posterior predictive is itself, again, often intractable. 

Most work on scaling Bayesian inference has focused on tackling the first problem of posterior inference: how can we sample or approximate $p(\theta | D)$? The two major approaches are (1) Markov Chain Monte Carlo methods that aim to form a sampling chain that generates samples from the posterior, and (2) variational methods that search for a parametric distribution that is as close as possible to the true Bayesian posterior. % ~\citep{kevin}. 
Given infinite time to run Monte Carlo, or an infinitely flexible variational family, we could recover samples from the Bayesian posterior. Unfortunately, these approaches only address the first of our two problems.  We still need to marginalize out this distribution to form the posterior predictive and inaccuracies in the approximate posterior could propagate to inaccuracies in the approximate posterior predictive.

Why spend so much time, compute, and energy to estimate the posterior over parameters if our primary goal is to form accurate predictions? 
Can we shortcut the two-step process and target the predictive distribution directly?
It turns out we can.

\section{Variational Prediction}

What we want is a direct variational predictive distribution that won't require marginalization. What we need is a principle to guide our search.  
What we'll do is start by considering two different ways to describe the world. 

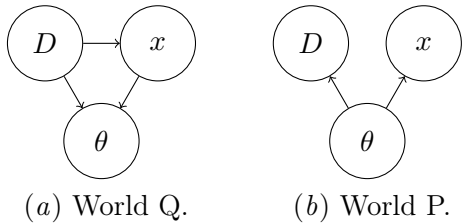
\begin{figure}[!ht]
\centering
{%
 \subfigure[World Q.]{\label{fig:worldq}%
   \begin{tikzpicture}[minimum size=1cm]
\def\r{0.5 cm}
\node [draw, circle] (D) at (0, 0) {$D$};
\node [draw, circle] (X) at (3*\r, 0) {$x$};
\node [draw, circle] (Theta) at (1.5*\r, -2.6*\r) {$\theta$};
\draw [->] (D) -- (Theta);
\draw [->] (D) -- (X);
\draw [->] (X) -- (Theta);
\end{tikzpicture}
}
}\qquad {%
\subfigure[World P.]{\label{fig:worldp}%
   \begin{tikzpicture}[minimum size=1cm]
\def\r{0.5 cm}
\node [draw, circle] (D) at (0, 0) {$D$};
\node [draw, circle] (X) at (3*\r, 0) {$x$};
\node [draw, circle] (Theta) at (1.5*\r, -2.6*\r) {$\theta$};
\draw [->] (Theta) -- (D);
\draw [->] (Theta) -- (X);
\end{tikzpicture}
}}
    \caption{The graphical models under consideration.
     $q(D,x,\theta) = q(D)q(x|D)q(\theta|x,D)$ and 
    $p(D,x,\theta) = p(D|\theta)p(x|\theta)p(\theta)$
    }
    \label{fig:worlds}
\end{figure}

In World $P$, the Bayesian world, we assume that we start by drawing a parameter value $\theta$ from some prior $p(\theta)$.  This parameter value then generates not only the data we observe $p(D|\theta)$ but also generates any future data (our predictions) $p(x|\theta)$.  Alternatively, in the real world, World $Q$, we start by observing data $D$ drawn from some process outside of our control $q(D)$. We want to create a process by which we could use that data to directly make predictions, $q(x|D)$.  Without any additional assumptions we allow for $\theta$ to depend on \emph{both} of these: $q(\theta|x,D)$.

To learn our variational predictive model, $q(x|D)$, we will align $Q$ and $P$ by minimizing the conditional KL:\footnote{All $\langle \cdot \rangle$ brackets in this work are expectations with respect to the full $q$ distribution. $\langle \cdot \rangle \equiv \langle \cdot \rangle_q \equiv \mathbb{E}_{q}[\cdot]$}
\begin{equation}
\left\langle \log \frac{q(x,\theta|D)}{p(x,\theta|D)} \right\rangle 
%\int dx\, d\theta\, q(x,\theta|D) \log \frac{q(x,\theta|D)}{p(x,\theta|D)} 
% \mathcal J &= 
% \left\langle \log \frac{q(x,\theta|D)}{p(x,\theta|D)} \right\rangle 
= \left\langle \log \frac{q(x|D)q(\theta|x,D)}{p(x|\theta)p(\theta|D)}  \right\rangle 
= \left\langle \log \frac{q(x|D) q(\theta|x,D)}{p(x|\theta) p(D|\theta) p(\theta)} \right\rangle + \log p(D) \geq 0 . \label{eqn:condklbound}
\end{equation}
%\Cref{eqn:condklbound} 
This establishes the Variational Prediction (VP) loss,
\begin{equation}
    \mathcal J \equiv \left\langle \log \frac{q(x|D) q(\theta|x,D)}{p(x|\theta) p(D|\theta) p(\theta)} \right\rangle ,
    \label{eqn:vploss}
\end{equation}
as a variational upper bound on the negative Bayesian marginal likelihood $(-\log p(D))$. 

Furthermore, the joint conditional KL (\cref{eqn:condklbound}) is also a variational upper bound on the KL divergence of our approximate predictive $q(x|D)$ to the Bayesian posterior predictive $p(x|D)$:
\begin{equation}
  \left\langle \log \frac{q(x,\theta|D)}{p(x,\theta|D)} \right\rangle =
 \left\langle \log \frac{q(x|D)}{p(x|D)} \right\rangle + \left\langle \log \frac{q(\theta|x,D)}{p(\theta|x,D)} \right\rangle
 \geq \left\langle \log \frac{q(x|D)}{p(x|D)} \right\rangle \geq 0.
 \label{eqn:predbound}
\end{equation}
Since \cref{eqn:condklbound} and \cref{eqn:vploss} differ only by the marginal likelihood, $\log p(D)$, if we fix the Bayesian model, $P$, this bound (\cref{eqn:predbound}) helps ensure that our approximate-predictive $q(x|D)$ will approximate the true Bayesian posterior predictive. The tightness of this bound is controlled by how well $q(\theta|x,D)$ matches $p(\theta|x,D)$, an \emph{augmented posterior}, 
i.e. the posterior you would obtain if you had observed an augmented dataset consisting of the original data and an additional observation at $x$. 
For some additional, independent insight, see \cref{app:candidate}.
% Having identified our VP loss, let's try to better understand how to use it.
% 
% In order to specify our loss, we require a Bayesian statistical model $p(x|\theta)$ and prior $p(\theta)$. Additionally, we need two variational approximations, $q(x|D)$ and $q(\theta|x,D)$. $q(x|D)$ is the variational predictive distribution we were after, 
% will form our variational approximation of the posterior predictive,
% while $q(\theta|x,D)$ acts like an \emph{augmented posterior}, 

Notice that \cref{eqn:predbound} ensures that our loss is a valid variational bound on the KL divergence between our variational predictive distribution, $q(x|D)$, and the true Bayesian predictive, $p(x|D)$, even if our variational augmented posterior is imperfect, our model or variational predictive is misspecified, or we have a finite dataset.

\subsection{Conditioning}

Extending the objective to conditional densities (e.g. regression rather than density estimation) is straightforward:
\begin{equation}
    \mathcal J_| = \left\langle \log \frac{q(y|x,D)q(x|D) q(\theta|y,x,D)}{p(y|x,\theta)p(x) p(D|\theta) p(\theta)} \right\rangle_{q} + \log p(D) \geq 
     \left\langle \log \frac{q(y|x,D)}{p(y|x,D)} \right\rangle_{q} .
     \label{eqn:vpcond}
\end{equation}
This clearly bounds the KL divergence between a variational conditional posterior, $q(y|x,D)$, and the true Bayesian conditional posterior, $p(y|x,D)$.
Interestingly, this KL divergence is minimized with respect to the $q$ distribution, namely and importantly this includes a new $q(x|D)$, a distribution used to generate our synthetic input points and under our control.  This means that we can choose which points to evaluate our predictive model at during training. In particular, for a classification or regression setting, if we have access to test-set inputs or unlabelled data, we could directly focus our efforts on learning a variational predictive that matched the true Bayesian predictive \emph{on those points}.

\subsection{Implicit Variational Augmented Posteriors}

Operationally, we follow the procedure illustrated in \cref{app:pseudo}. While minimizing the objective we (1) synthesize a new data-point $x$ from our variational predictive distribution $q(x|D)$, (2) compute and sample a parameter value from some variational augmented posterior $\theta \sim q(\theta|x,D)$, and (3) score that synthetic data-point and parameter value according to their sources $q(x|D), q(\theta|x,D)$ and the Bayesian model $p(x|\theta)p(D|\theta)p(\theta)$.  In contrast to most other forms of inference, our predictive models outputs are judged during the training process.  This allows the VP method to scrutinize the predictive model off the data manifold.

% The objective rewards \emph{synthetic} datapoints, $x$, and parameter settings, $\theta$, that are consistent with the data according to the statistical model $p$. That is, we are trying to create a data generator, $q(x|D)$, that generates data consistent with our Bayesian model, $p(D,\theta)$.  We additionally require a variational approximation, $q(\theta|x,D)$, to the Bayesian posterior that would be induced had we observed the synthetic data alongside our original data, $p(\theta|x,D)$.

% \fixme{Maybe as a common question / point of comparison, here we could describe an alternate procedure: first, fit a variational posterior directly. Then, freezing that variational posterior, fit a variational predictive. You could do this with a Bayesian dark knowledge-y approach where you combine varaitional posterior $q(\theta |x)$ and true likelihood $p(x | \theta)$}

In general, specifying an augmented posterior means building a \emph{conditional} posterior approximation.  For large-scale problems this naively creates an new intractable task.
%This could, in general, be quiet expensive.  In large-scale settings, there may be millions of parameters, so a mean field variational approximation would already constitute a couple million parameters by itself, in order to make those parameters conditioned on new data would naively mean using a neural network that would have a couple-million dimensional output.  This is much too expensive to build in practice so we need some more efficient way to build expressive conditional augmented posteriors. \fixme{trim paragraph a bunch}
%
Inspired by MAML \citep{maml}, in the toy experiments below, we've defined the augmented posterior implicitly
as a single gradient update of an approximate (unconditional) posterior:
\begin{align}
  q(\theta| y,x,D) &= q(\theta' |D) \\
  \theta' &= \theta - \lambda \nabla \left\langle -\beta \log p(y|x,\theta) + \log\frac{q(\theta|D)}{p(\theta)} \right\rangle_{q(\theta|D)} .
  \label{eqn:maml}
\end{align}
This requires specifying an approximate posterior $q(\theta|D)$ (at the same cost as in variational Bayes), plus two additional parameters: $\lambda$, a learning rate for the update, and $\beta$, an effective inverse temperature for the ELBO used to update the approximate posterior.  This $\beta$ sets the effective number of observations the synthetic point is considered equivalent to during the posterior update.

\section{Toy Example}

Let's demonstrate the method on a simple toy example.  
%
% To highlight the strengths of the method, we want an example of a problem for which the exact Bayesian posterior predictive is decent, but for which a simple variational approximation is lacking.  In particular, we're interesting in establishing whether we can use the variational prediction objective to learn a predictive distribution that works even when using a simple approximate posterior, where if we first learned a simple approximate posterior and integrated it out we'd end up with a worse predictive distribution.  
% \fixme{could cut the apologies for the choice of toy or move it later down.}
Consider a two-parameter sinusoidal curve-fitting problem: 
\begin{align}
    x &\sim \mathcal U(0,1) \\
    y &\sim \mathcal N( \mu(x), 1 ) \\
    \mu(x) &= \sin\left( 2 \pi f x + \phi \right) .
\end{align}
The $x$ values are drawn uniformly on the unit interval.  The $y$ values follow a sinsusoidal curve with frequency, $f$, and phase, $\phi$.  The observational model is Gaussian with unit variance.  For our target distribution we'll set $f=1, \phi=1$ and sample 8 data-points to serve as our dataset as shown in {\color{Cblue!70!black} blue} in \cref{fig:solution} below.  The predictive distributions are shown in {\color{Corange!70!black} orange}.

%\subsection{Bayesian Inference}
\begin{figure}[!htb]
    \centering
    \includegraphics[width=\textwidth]{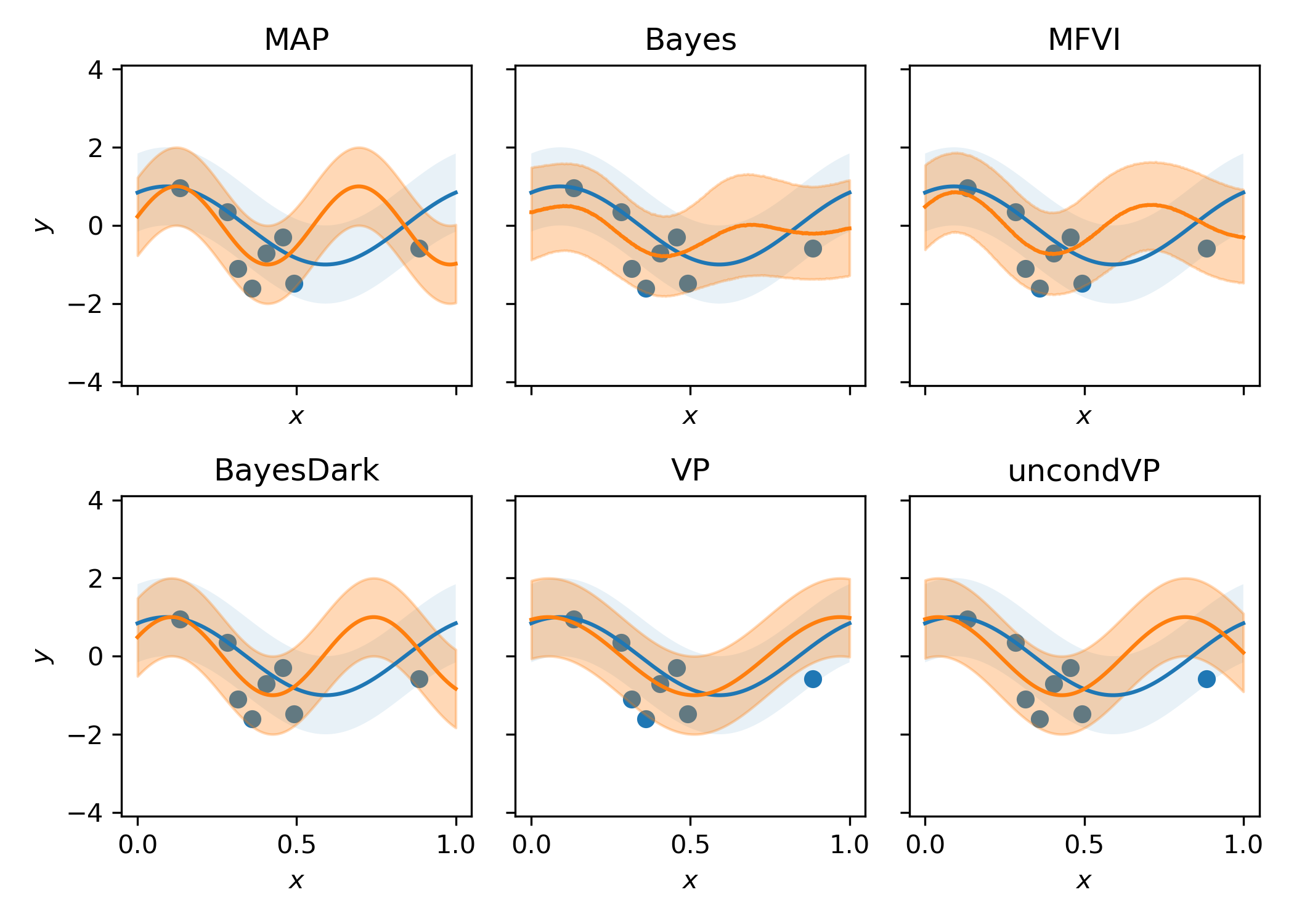}
    \caption{Six different predictive distributions learned for the toy example. In each panel, the true distribution and data are shown in {\color{Cblue!70!black} blue}, the fit predictive distribution is shown in {\color{Corange!70!black} orange}.
    % From left to right, there is:
    % (1) the Maximum a posteriori (MAP) predictive distribution, (2) the exact Bayesian posterior predictive (Bayes), (3) the predictive distribution obtained by marginalizing out a Mean Field Variational Inference (MFVI) approximate posterior, (4) the Bayesian Dark Knowledge (BayesDark) distilled predictive distribution, (5) the proposed Variational Predictive (VP) method and (6) a baseline version of Variational Prediction that uses an unconditional augmented posterior, i.e. the same approximate posterior used for the other methods.
    }
    \label{fig:solution}
    \label{fig:solutions}
\end{figure}

We compared six different inference techniques:
(1) Maximum a posteriori (MAP) estimation of the prediction distribution,
 (2) the exact Bayesian posterior predictive (Bayes), with a $\mathcal N(0,16)$ prior on both $\log f$ and $\phi$,
 (3) the predictive distribution obtained by marginalizing out a Mean Field Variational Inference (MFVI) approximate posterior, using a factorized Gaussian,
 (4) the Bayesian Dark Knowledge (BayesDark) \citep{bayesdark} distilled predictive distribution, 
 (5) the proposed Variational Predictive (VP) method, and
 (6) a baseline version of Variational Prediction that uses an unconditional augmented posterior (uncondVP), i.e. the same approximate posterior used for the other methods. 

%\subsection{Mean Field Variational Inference}

\begin{figure}[!htb]
    \centering
    \includegraphics[width=\textwidth]{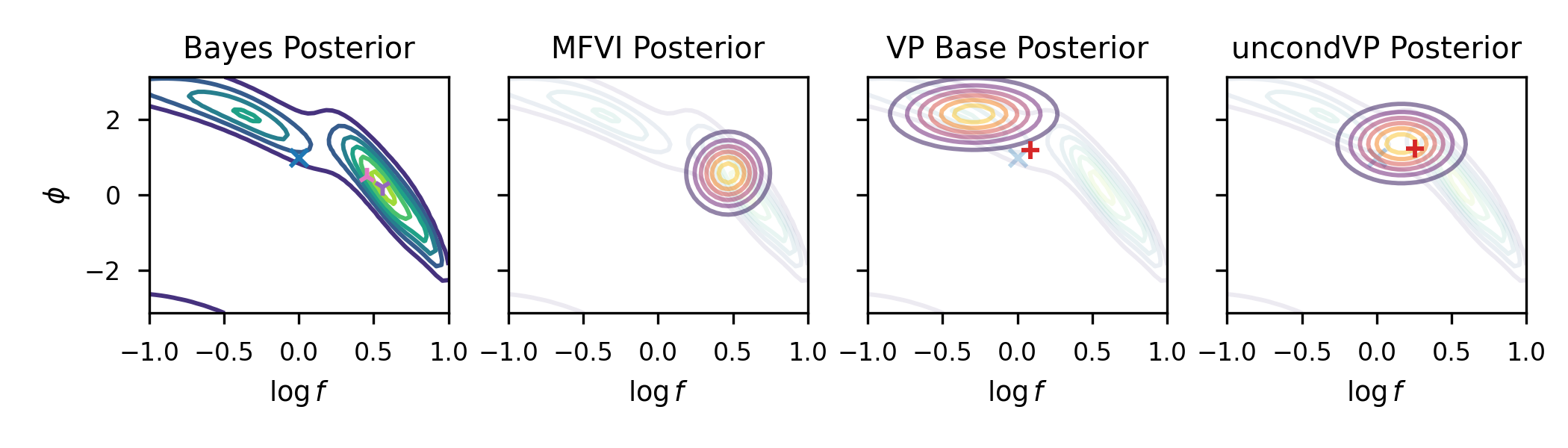}
    \caption{The exact Bayesian posterior for the data shown in {\color{Cblue!70!black} blue} in \cref{fig:solution} is shown on the left. The $\boldsymbol{\color{Cblue} \times}$ marks the true parameter values. The $\boldsymbol{\color{Cpurple} \Ydown}$ marks the MAP parameters, $\boldsymbol{\color{Cpink} \Yup}$ marks the BayesDark parameters. The second and last column show the MFVI and uncondVP learned approximate posteriors, respectively. For the VP example, we learned a conditional posterior which is hard to visualize, shown is the base unconditional posterior that is modified with MAML. 
    For the VP and uncondVP solutions the corresponding learned predictive distribution parameters are indicated by $\boldsymbol{\color{Cred} +}$.
    }
    \label{fig:posteriors}
\end{figure}

% combined figure side by side if pressed for space.
% \begin{figure}[!htb]
% \centering
% %{%
%  \subfigure[Predictive Distributions.]{\label{fig:solutions2}%
%     \includegraphics[width=0.55\textwidth]{images/vpp-solution.png}
%   }
% % }\qquad {%
% %
% \subfigure[Posteriors.]{\label{fig:posteriors2}%
%     \includegraphics[width=0.40\textwidth]{images/posteriors3.png}
% }
% %}
%     \caption{Combined figure.}
%     \label{fig:combined}
% \end{figure}

In \cref{fig:posteriors} we visualize the solutions in parameter space. Notice that the true Bayesian posterior (left panel) is bimodal and neither mode is particularly well aligned with the true parameters ($\boldsymbol{\color{Cblue} \times}$).  It's corresponding prediction distribution cannot be represented by a single frequency and phase. The MAP predictive distribution (top-left of \cref{fig:solutions}) uses a single point-estimate, the global maximum of the true Bayesian posterior ($\boldsymbol{\color{Cpurple} \Ydown}$ in \cref{fig:posteriors}).
%Marginalizing out this posterior, the exact Bayesian posterior predictive shown in the top-center panel of \cref{fig:solution} in orange shows the effect of both frequencies.
The learned MFVI posterior is shown in the second column of \cref{fig:posteriors}.  As is usually the case, variational Bayesian inference learns an approximate posterior that aims to minimize the KL divergence from the approximate posterior to the true, which leads to a mode-seeking behavior and the approximate posterior, here, concentrates at the higher frequency mode. This means that the posterior predictive formed when marginalizing out this approximate posterior has predominately the wrong frequency, shown in the top-right of \cref{fig:solution}.

To learn direct (marginalization-free) variational predictive models, we first replicate the Bayesian Dark Knowledge technique of \citet{bayesdark}, which attempts to \emph{distill} the true Bayesian posterior predictive distribution into a parametric $q(y|x,D)$, minimizing $KL[p(y|x,D) | q(y|x,D)]$, the opposite KL compared with the VP objective (\cref{eqn:vpcond}).  This method requires exact samples from the true Bayesian posterior predictive. For the learned predictive model, $q(y|x,D)$, we simply used the true likelihood $q(y|x,D) = p(y|x,\theta)$ with learned $\log f$ and $\phi$ parameters.  The resulting predictive model is shown on the bottom left of \cref{fig:solutions} with the learned parameters denoted with $\boldsymbol{\color{Cpink} \Yup}$ in \cref{fig:posteriors}. Notice that these parameters are at the center of the MFVI approximate posterior.

For the conditional VP method (\cref{eqn:vpcond}), we used a uniform $\mathcal U(0,1)$ distribution for the synthetic $x$ distribution $q(x|D)$.  We used a parametric copy of the likelihood for the predictive model, $q(y|x,D)$, as with BayesDark, and used the MAML style augmented posterior (\cref{eqn:maml}) acting on the factorized Gaussian posterior of the MFVI method.

% In general, specifying an augmented posterior means building a conditional posterior approximation.  This could in general be quiet expensive.  In large scale settings, there may be millions of parameters, so a mean field variational approximation would already constitute a couple million parameters by itself, in order to make those parameters conditioned on new data would naively mean using a neural network that would have a couple-million dimensional output.  This is much too expensive to build in practice so we need some more efficient way to build expressive conditional augmented posteriors.

% Instead, inspired by MAML \citep{maml}, here we've defined the augmented posterior implicitly
% as a single gradient update of an approximate (unconditional) posterior:
% \begin{align}
%   q(\theta| y,x,D) &= q(\theta' |D) \\
%   \theta' &= \theta - \lambda \nabla \left\langle -\beta \log p(y|x,\theta) + \log\frac{q(\theta|D)}{p(\theta)} \right\rangle_{q(\theta|D)} .
% \end{align}
% This requires specifying an approximate posterior $q(\theta|D)$ (at the same cost as in variational Bayes), plus only two additional parameters: $\lambda$, a learning rate for the update and $\beta$, an effective inverse temperature for the ELBO used to update the approximate posterior.  This $\beta$ sets the effective number of observations the synthetic point is considered equivalent to during the posterior update.  

With these choices in place, the learned predictive model in the bottom-left panel of \cref{fig:solution} is a nice fit to the true distribution.  We emphasize that this predictive model doesn't require any marginalization and is defined by a point estimate for the parameters of the likelihood.  The specific parameter values found are shown by the $\boldsymbol{\color{Cred}+}$ in \cref{fig:posteriors}, along with the learned base augmented posterior, i.e. the learned approximate posterior that gets updated by the synthetic point.  The learned effective-learning-rate was $\lambda = 0.004$ and the learned effective-inverse-temperature was $\beta = 12.8$ in this example.

In all of the other inferential methods, the learned predictive distribution was only ever scored on the data itself, it was never tasked with generating predictions. 
Meanwhile, in the VP method we are using the predictive model at training time to generate new synthetic data that also must be explained by the Bayesian model.  We suspect that it's this enforced sense of in internal consistency that improves the VP predictive model.

To isolate the effect and utility of the MAML-style conditioning in the augmented posterior, lastly we reran the Variational Prediction method but used an unconditional augmented posterior (uncondVP), i.e. the same mean-field approximate posterior that was used for the MFVI experiment. Now, the learned approximate posterior is tasked with explaining not only the original data, but also a single synthetic draw from the predictive model.  This isn't equivalent to the MFVI method because the approximate posterior sees the additional point; and it's not equivalent to the VP method because the posterior no longer differs under distinct draws from the predictive model. Removing the conditioning leads to a worse predictive distribution as shown in the bottom right of \cref{fig:solutions}. 

\section{Conclusion}

The toy model illustrates that Variational prediction can learn cheap but good predictive models at little additional cost beyond variational Bayes. However, we've encountered issues trying to scale the VP objective up to larger problems, mainly due to variance in the loss which makes convergence difficult. We hope to resolve these issues in future work, perhaps with techniques such as in \citet{sticking, void, mamlpp}.

In this paper, we've introduced the VP method. 
VP is a new inferential technique for learning variational approximations to Bayesian posterior predictive distributions
that doesn't require 
(1) the posterior predictive distribution itself,
(2) the exact posterior distribution,
(3) exact samples from the posterior,
(4) or any test time marginalization. 
We are excited to see if this method can be shown to be workable on larger-scale problems.% and already believe it offers a fresh perspective.

%\begin{listing}[!ht]
%\begin{minted}{python}
%def foo(x):
  %return x
%\end{minted}
%\caption{This is a caption.}
%\end{listing}

%\end{document}

%\acks{Acknowledgements go here.}
%\acks{James Harrison, Joshua V Dillon, Sergey Ioffe, Jascha Sohl Dickstein, Kevin Murphy, Ian Fischer}

\bibliography{bib}

\begin{thebibliography}{13}
\providecommand{\natexlab}[1]{#1}
\providecommand{\url}[1]{\texttt{#1}}
\expandafter\ifx\csname urlstyle\endcsname\relax
  \providecommand{\doi}[1]{doi: #1}\else
  \providecommand{\doi}{doi: \begingroup \urlstyle{rm}\Url}\fi

\bibitem[Agakov and Barber(2004)]{agakov2004auxiliary}
Felix~V Agakov and David Barber.
\newblock An auxiliary variational method.
\newblock In \emph{Neural Information Processing: 11th International
  Conference, ICONIP 2004, Calcutta, India, November 22-25, 2004. Proceedings
  11}, pages 561--566. Springer, 2004.

\bibitem[Aitchison(1975)]{aitchison1975goodness}
James Aitchison.
\newblock Goodness of prediction fit.
\newblock \emph{Biometrika}, 62\penalty0 (3):\penalty0 547--554, 1975.

\bibitem[Antoniou et~al.(2019)Antoniou, Edwards, and Storkey]{mamlpp}
Antreas Antoniou, Harrison Edwards, and Amos Storkey.
\newblock How to train your maml, 2019.

\bibitem[Balan et~al.(2015)Balan, Rathod, Murphy, and Welling]{bayesdark}
Anoop~Korattikara Balan, Vivek Rathod, Kevin~P Murphy, and Max Welling.
\newblock Bayesian dark knowledge.
\newblock In \emph{Advances in Neural Information Processing Systems}, pages
  3438--3446, 2015.

\bibitem[Besag(1989)]{candidate}
Julian Besag.
\newblock A candidate's formula: A curious result in bayesian prediction.
\newblock \emph{Biometrika}, 76\penalty0 (1):\penalty0 183--183, 1989.

\bibitem[Boisbunon and Maruyama(2014)]{boisbunon2014inadmissibility}
Aur{\'e}lie Boisbunon and Yuzo Maruyama.
\newblock Inadmissibility of the best equivariant predictive density in the
  unknown variance case.
\newblock \emph{Biometrika}, 101\penalty0 (3):\penalty0 733--740, 2014.

\bibitem[Finn et~al.(2017)Finn, Abbeel, and Levine]{maml}
Chelsea Finn, Pieter Abbeel, and Sergey Levine.
\newblock Model-agnostic meta-learning for fast adaptation of deep networks,
  2017.

\bibitem[George et~al.(2006)George, Liang, and Xu]{george2006improved}
Edward~I George, Feng Liang, and Xinyi Xu.
\newblock Improved minimax predictive densities under kullback-leibler loss.
\newblock \emph{The Annals of Statistics}, pages 78--91, 2006.

\bibitem[Grathwohl et~al.(2018)Grathwohl, Choi, Wu, Roeder, and Duvenaud]{void}
Will Grathwohl, Dami Choi, Yuhuai Wu, Geoffrey Roeder, and David Duvenaud.
\newblock Backpropagation through the void: Optimizing control variates for
  black-box gradient estimation, 2018.

\bibitem[Liang and Barron(2004)]{liang2004exact}
Feng Liang and Andrew Barron.
\newblock Exact minimax strategies for predictive density estimation, data
  compression, and model selection.
\newblock \emph{IEEE Transactions on Information Theory}, 50\penalty0
  (11):\penalty0 2708--2726, 2004.

\bibitem[Roeder et~al.(2017)Roeder, Wu, and Duvenaud]{sticking}
Geoffrey Roeder, Yuhuai Wu, and David Duvenaud.
\newblock Sticking the landing: Simple, lower-variance gradient estimators for
  variational inference, 2017.

\bibitem[Snelson and Ghahramani(2005)]{snelson2005compact}
Edward Snelson and Zoubin Ghahramani.
\newblock Compact approximations to bayesian predictive distributions.
\newblock In \emph{Proceedings of the 22nd international conference on Machine
  learning}, pages 840--847, 2005.

\bibitem[Welling and Teh(2011)]{sgld}
Max Welling and Yee~W Teh.
\newblock Bayesian learning via stochastic gradient langevin dynamics.
\newblock In \emph{Proceedings of the 28th international conference on machine
  learning (ICML-11)}, pages 681--688, 2011.

\end{thebibliography}

\appendix

\section{Candidate's Formula}
\label{app:candidate}

In a single-page paper,~\citet{candidate} shares a curious result that
appeared on a candidate's final exam:
\begin{equation}
    \label{eqn:student}
    p(x|D) = \frac{p(x|\theta) p(\theta|D)}{p(\theta|x,D)}.
\end{equation}
Remarkably the right hand side holds for any $\theta$ if the $p$'s are all taken to be consistent with some Bayesian model $p(x|\theta)p(\theta)$. The Bayesian posterior predictive distribution can be found by relating the likelihood, posterior and an \emph{augmented posterior}, namely the posterior you would compute if you observed not only the data $(D)$,
but also the prediction $x$ as an additional datapoint. 

The proof is straightforward: simply factorize the joint distribution $p(x,\theta,D)$ out two ways:
$p(\theta,x,D)= p(\theta|x,D) p(x|D) = p(x|\theta) p(\theta|D)$
and notice that because of the Markov chain 
$D \to \theta \to x$ we have that $p(x|\theta, D) = p(x|\theta)$. $\blacksquare$

As this demonstrates, if we knew the exact augmented posterior, we would be able to determine the exact posterior predictive without needing to explicitly marginalize.  Our variational approach instead steers a variational approximation to the posterior predictive by using a variational approximation to the augmented posterior.

\section{Pseudo-code}
\label{app:pseudo}

\begin{listing}[!ht]
\begin{minted}[fontsize=\footnotesize]{python}
def variational_prediction(predictive_model, aug_posterior, likelihood, prior, data):
    # first we sample from the predictive model
    x = predictive_model.sample()
    # then we compute the approximate posterior induced by that sample
    approx_posterior = aug_posterior(x)
    # now we proceed as we would normally with and ELBO by sampling a parameter value
    theta = approx_posterior.sample()
    # and then computing all of the relevant log probabilities.
    return (predictive_model.log_prob(x) + approx_posterior.log_prob(theta) 
            - likelihood.log_prob(x) - likelihood.log_prob(data) - prior.log_prob(theta))
\end{minted}
\caption{A (pseudo-)python implementation of the variational prediction loss.}
\label{lst:vpp}
\end{listing}

\section{Related Work}
\label{app:related}
% 
% There are nice things that can be said the convergence of MCMC methods \fixme{CITATION NEEDED} as well as variational Bayes \citep{mispecified-variational}.
% 
\citet{aitchison1975goodness} introduced the distinction between \emph{estimative} procedures that use point estimates of parameters and \emph{predictive} approaches that involved marginalizing over parameter distributions.  In that work, Aitchison showed that the Bayesian posterior predictive distribution is optimal in terms of the average Kullback-Leibler divergence from the true distribution as well as how some predictive distributions can always outperform standard estimative approaches like Maximum Likelihood by this same metric.

Follow-up work~\citep{george2006improved,liang2004exact,boisbunon2014inadmissibility} have extended and further analyzed these types of scenarios.  However, in these cases the predictive density generated requires the form of marginalization we aimed to avoid in this work.  Here we are interested in defining a type of \emph{estimative} procedure that can claim to more directly target predictive performance.  We did so by use of an auxillary variational method to generate our bounds, akin to~\citet{agakov2004auxiliary}. 

Both \citet{snelson2005compact} and \citet{bayesdark} attempt to learn compact \emph{estimative} representations of the Bayesian posterior predictive, though targeting the opposite KL divergence as done here.  They aim to learn a model for the predictive $q(x|D)$ that is as close as possible to the true posterior predictive $p(x|D)$ as measured by $\left\langle \log \frac{p(x|D)}{q(x|D)} \right\rangle_{p(x|D)}$. This requires being able to generate samples from the true Bayesian posterior predictive.  In~\citet{bayesdark}, Stochastic Gradient Langevin Dynamics~\citep{sgld} was used to generate these samples.  These samples were then distilled into a compact distribution. In our work we don't need to presuppose we can generate exact samples from the true Bayesian posterior and have more freedom to leverage tractable variational approximations to the augmented posterior while maintaining our bounds.

% 
% \section{Experimental Details}
% \label{app:details}
% 
% To further elaborate on the experiments on the toy model, our data generating process took the form:
% \begin{align}
%     x &\sim \mathcal U(0, 1) \\
%     y &\sim \mathcal N(\mu(x|f,\phi), 1) \\
%     \mu(x|a,\phi) &= \sin\left(2 \pi f x + \phi \right)
% \end{align}
% \begin{equation}
%     p(x|f, \phi) = \mathcal N(\mu(x), 1) = \frac{1}{\sqrt{2\pi}} e^{-\frac{(x-\sin(2\pi f x + \phi)))^2}{2}}
% \end{equation}

% \fixme{todo}
% \subfile{tex/appendices}

\end{document}